\documentclass[sigconf]{acmart}

\usepackage{enumitem}
\usepackage{multirow}
\usepackage{float}
\AtBeginDocument{%
  }

\setcopyright{none}
\copyrightyear{}
\acmYear{2027}
\acmDOI{}
\acmISBN{}
\acmConference[Conference acronym 'XX]{Make sure to enter the correct
  conference title from your rights confirmation email}{June 03--05,
  2018}{Woodstock, NY}

\makeatletter
\providecommand{\@currentauthors}{}
\providecommand{\@currentaffiliation}{}
\makeatother

\begin{document}

\title{Utility-Aware Generative Recommendation: Utility-Enriched Semantic Codes and Utility-Guided Decoding}


\author{Yang Hu}
\authornote{These authors contributed equally to this work.}
\authornote{Yang Hu is an intern at Kuaishou. He is now at
Peking University.}
\email{huyang13@kuaishou.com}

\author{Duo Wang}
\authornotemark[1]
\email{wangduo06@kuaishou.com}

\author{Ning Li}
\authornote{Corresponding author.}
\email{lining10@kuaishou.com}

\author{Kang Li}
\email{likang14@kuaishou.com}

\author{Junda She}
\email{shejunda03@kuaishou.com}

\author{Chengcheng Guo}
\email{guochengcheng03@kuaishou.com}

\author{Kuo Cai}
\email{caikuo@kuaishou.com}

\author{Bin Liu}
\email{liubin21@kuaishou.com}

\author{Youhua Liu}
\email{liuyouhua@kuaishou.com}

\author{Xiaoshuang Chen}
\email{chenxiaoshuang@kuaishou.com}

\affiliation{
  \institution{Kuaishou Technology}
  \city{Beijing}
  \country{China}
}

\renewcommand{\shortauthors}{Hu et al.}

\makeatletter
\renewcommand{\@mkauthors@iii}{%
  \global\setbox\mktitle@bx=\vbox{%
    \unvbox\mktitle@bx
    \par\medskip
    \centering

    {\normalfont\fontsize{12}{10}\selectfont
      Yang Hu$^{*,\dagger}$, Duo Wang$^{*}$, Ning Li$^{\ddagger}$,
      Kang Li, Junda She, Chengcheng Guo,\\
       Kuo Cai, Bin Liu,
      Youhua Liu, Xiaoshuang Chen\par
    }

    \smallskip
    {\normalfont\fontsize{10}{10}\selectfont
      Kuaishou Technology, Beijing, China\par
    }

    \smallskip
    {\normalfont\fontsize{10}{10}\selectfont
      \{huyang13, wangduo06, lining10, likang14, shejunda03, guochengcheng03,\\
       caikuo, liubin21, liuyouhua,
      chenxiaoshuang\}@kuaishou.com\par
    }

    \bigskip
  }%
}
\makeatother


\begin{abstract}
Generative recommendation has been widely adopted in large-scale industrial
recommendation systems, where multimodal item embeddings are quantized into
offline semantic identifiers (SIDs) and items are retrieved by
autoregressively generating their SIDs. 
However, conventional SID construction often underrepresents item utility reflected in post-exposure behaviors such as shares, long views, and video completion, making semantically similar items with different utility insufficiently distinguishable while struggling to accommodate rapidly changing utility signals.
Moreover, generation likelihood may be misaligned with downstream utility, causing high-utility candidates to be prematurely pruned during beam search before post-generation reranking or training-time reward alignment can effectively compensate.
To address these limitations, we propose \textbf{UAGR}, which aligns both SID
construction and decoding with downstream multi-objective utility. 
Our Utility-Enriched Semantic Codes preserve the original semantic information while incorporating item-level utility without increasing SID length or vocabulary size, supporting incremental utility updates with no additional tokenization or sequence-decoding cost.
Meanwhile, our Utility-Guided Decoding estimates candidate utility by combining predictions of multiple post-exposure outcomes and incorporates these estimates into beam scoring, reducing the risk of prematurely pruning high-utility paths while enabling flexible utility control without interfering with generative backbone training.
Extensive offline experiments and online A/B tests under real-world traffic demonstrate the effectiveness and deployability of UAGR, which has been fully deployed on Kuaishou and increases video watch time by \textbf{0.206\%} online.
\end{abstract}




\maketitle

\section{Introduction}
\label{sec:introduction}
Generative recommendation has emerged as a practical paradigm for large-scale
industrial recommendation and has been deployed across short-video,
advertising, live-streaming, and local-service platforms~\cite{deng2025onerec,xue2026gr4ad,wang2026onelive,wei2026oneloc}.
Unlike conventional retrieval systems that score items independently,
identifier-based generative recommenders represent each item as a semantic identifier (SID) composed of discrete tokens and autoregressively generate candidate SIDs conditioned on user interaction histories~\cite{rajput2023recommender,ju2025generativerecommendationsemanticids,Liu_2026}.
In a typical pipeline, multimodal item embeddings are discretized into
hierarchical SIDs using residual vector quantization~\cite{deng2025onerec,fu2026forge}, such as residual
K-means~\cite{chen2010residual} or
RQ-VAE~\cite{lee2022autoregressive}, and finite-width beam search is used to
explore the resulting discrete identifier space.

Despite this progress, a common training objective for generative recommenders is to maximize the likelihood of items in users’ interaction histories.
This objective captures relevance to past user behavior, but does not explicitly model the utility reflected in post-exposure outcomes such as long views, video completion, watch time, likes, or shares.
Candidates with similar generation probabilities can
therefore yield substantially different post-exposure outcomes, and the ordering induced by generation likelihood may differ from that induced by downstream utility.
This mismatch is particularly consequential under finite-width beam search: a path leading to a high-utility item may be pruned at an intermediate decoding step, thereby preventing the item from being recovered during subsequent decoding~\cite{sun2026historyconditionedjointprefixalignmentgenerative,wang2026understandingsemanticidsitem,sun2026semanticidsfinitebeamreachability}.

Recent studies have explored additional supervision and scoring mechanisms beyond standard next-token prediction to refine retrieved candidates or guide intermediate decoding.
GRank~\cite{sun2026grank} injects embeddings of target items and in-batch negatives during training to enhance candidate discrimination, retrieves candidates through vector search, and subsequently applies a lightweight target-aware ranker for refinement.
However, such post-retrieval reranking can only reorder retrieved candidates and cannot recover items excluded from the initial candidate set.
OneRec~\cite{deng2025onerec} unifies retrieval and ranking within a generative framework and further employs reinforcement learning to iteratively align SID generation with personalized preference signals.
However, such training-time sequence-level alignment optimizes the generation policy over complete SID sequences, while beam search makes irreversible pruning decisions over intermediate SID prefixes, and therefore does not necessarily ensure that high-utility candidates survive intermediate pruning.
RankGR~\cite{fu2026rankgr} aligns generation probabilities with preference orderings derived from behavioral feedback and uses a target-aware scorer trained with binary supervision to rescore candidate codewords during stepwise beam refinement.
However, its preference-based scoring does not explicitly estimate heterogeneous post-exposure outcomes that determine downstream multi-objective utility in production systems, potentially limiting its ability to identify and retain high-utility candidates during beam search.

Utility misalignment can also originate in SID construction.
Conventional semantic tokenizers commonly organize items according to semantic similarity through residual quantization~\cite{rajput2023recommender,ju2025generativerecommendationsemanticids}.
To move beyond this purely semantic organization, recent methods~\cite{zheng2024lcrec,wang2024letter,fu2026forge,NEURIPS2025_ee2841db} incorporate collaborative signals derived from user-item interactions through representation alignment, contrastive learning, or collaborative-similarity-guided tokenization.
Although these signals can improve item distinguishability and capture interaction-level relationships beyond content semantics, they neither explicitly characterize the downstream utility reflected in heterogeneous post-exposure outcomes nor necessarily preserve the original semantic structure when incorporated into the tokenizer.
Consequently, utility-relevant distinctions may still be insufficiently preserved in the discrete identifier space, while perturbations to the learned semantic structure can offset the benefits of collaborative enrichment and, in some cases, even degrade downstream performance relative to the original semantic tokenizer~\cite{dan2026crossstage}.
More importantly, collaborative signals are inherently time-varying and can shift substantially over short time scales, causing the information encoded in learned SIDs to become stale and limiting the practicality of such learning-based tokenizers in large-scale industrial deployment~\cite{Baikalov_2026}.

Motivated by these observations, we propose
\textbf{U}tility-\textbf{A}ware \textbf{G}enerative
\textbf{R}ecommendation (\textbf{UAGR}), which aligns both SID construction and
autoregressive decoding with downstream utility.
At the representation level, we introduce Utility-Enriched Semantic Codes 
(UE-SID), which improves item distinguishability through lookup-free quantization
at the final level and incorporates discretized item-level utility into the
final SID token, while preserving the hierarchical semantic structure.
This design explicitly encodes item-level utility in the discrete SID space
without increasing SID length or vocabulary size, simultaneously supporting
incremental utility updates without retraining the semantic quantizers.
At the decoding level, we propose Utility-Guided Decoding (UGD), which aggregates predictions of multiple post-exposure outcomes from intermediate decoding states into a configurable utility score, which is combined with generation likelihood during beam pruning to reduce the risk of prematurely discarding high-utility paths.
Meanwhile, during training, stop-gradient isolation prevents
utility-related gradients from updating the generative backbone,
thereby avoiding gradient interference between utility estimation
and next-token prediction, preserving the backbone's original
generative capability.
Extensive offline experiments and online A/B tests on Kuaishou,
a real-world recommendation platform serving hundreds of millions
of daily active users, validate the effectiveness and practicality
of UAGR, whose plug-and-play utility module supports flexible utility
configuration without interfering with the training of the online
generative model.
UAGR has been fully deployed on Kuaishou.

Our main contributions are summarized as follows:
\begin{itemize}[leftmargin=*, labelindent=0pt, labelsep=0.5em]
\item We propose UE-SID to explicitly encode item-level utility in fixed-length SIDs while preserving the hierarchical semantic structure, supporting incremental utility updates without increasing the vocabulary size.

\item We develop UGD to incorporate predictions of multiple post-exposure outcomes into beam scoring, enabling flexible utility control without interfering with generative backbone training.

\item We conduct extensive offline experiments and online A/B tests on
Kuaishou. The results demonstrate the effectiveness and
production feasibility of UAGR, which has been fully deployed on Kuaishou platforms.
\end{itemize}

\begin{figure*}[!t]
    \centering
    \makebox[\textwidth][c]{%
        \includegraphics[width=1.0\textwidth]{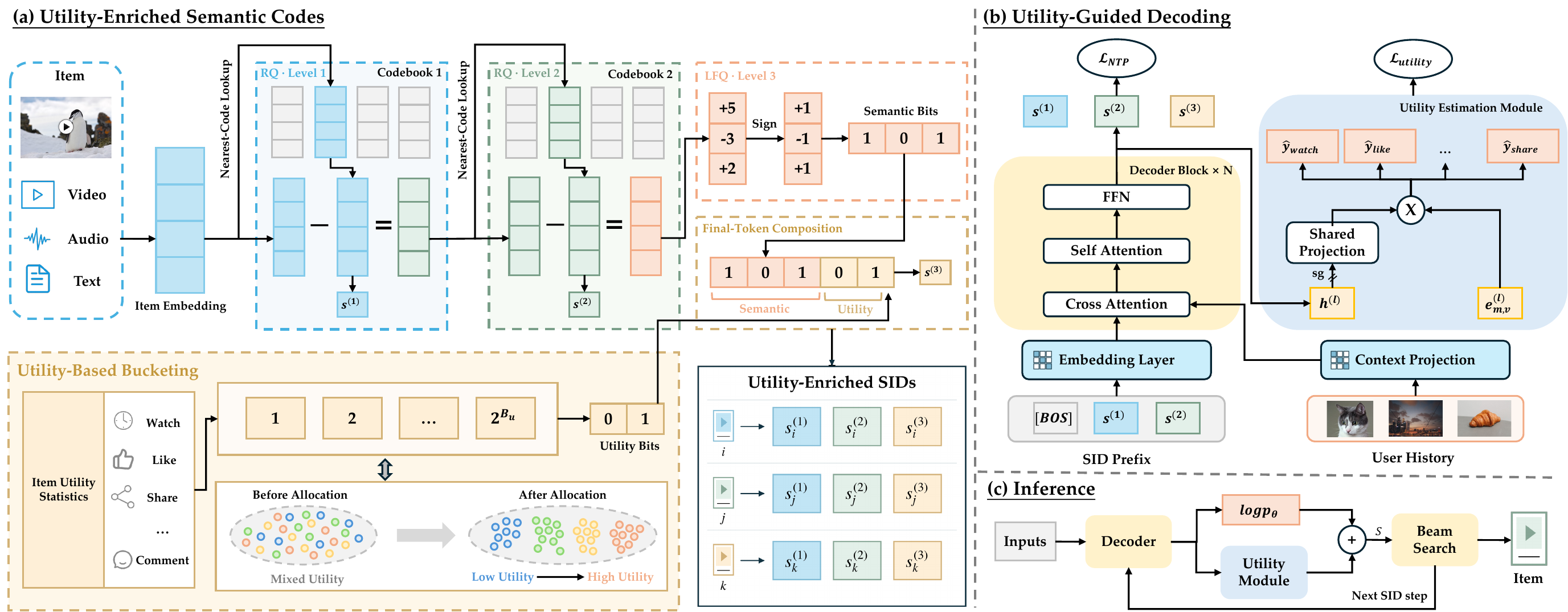}%
    }
    \caption{Overview of UAGR.
    (a) Utility-Enriched Semantic Codes encode item-level utility in the
    final SID token without increasing SID length or vocabulary size.
    (b) The utility estimation module, with stop-gradient isolation,
    predicts post-exposure feedback from decoder hidden states.
    (c) Utility-Guided Decoding combines cumulative generation
    log-likelihood with utility estimates to select candidate prefixes
    during beam search.}
    \Description{The UAGR framework consists of two complementary components.
    The first incorporates item-level utility into semantic identifiers.
    The second combines cumulative generation log-likelihood with predicted
    utility at each decoding step to select the prefixes retained by beam search.}
    \label{fig:framework}
\end{figure*}

\section{Related Work}
\label{sec:related-work}

\subsection{Generative Recommendation}
\label{sec:related_generative_recommendation}

Early generative recommenders adapt pretrained language models to
recommendation through fine-tuning. P5~\cite{geng2022p5} unifies multiple
recommendation tasks in a text-to-text framework, while
TALLRec~\cite{bao2023tallrec} aligns language models with recommendation
instructions and GPTRec~\cite{petrov2023gptrec} models personalized
recommendation as autoregressive generation. Although flexible, these methods
generate textual outputs that must be grounded to items in a large and
continuously changing catalog.

TIGER~\cite{rajput2023recommender} introduced identifier-based generative
recommendation by quantizing item embeddings into hierarchical semantic IDs
and autoregressively generating these IDs from user histories. This paradigm
has since been validated in industrial systems. OneRec~\cite{deng2025onerec}
unifies retrieval and ranking through generation and aligns the generator
with user preferences; GR4AD~\cite{xue2026gr4ad} co-designs item tokenization,
model learning, and online serving for advertising; and
OneLive~\cite{wang2026onelive} adapts generative recommendation to highly
dynamic live-streaming content. Related systems further extend the paradigm
to local services and advertising~\cite{wei2026oneloc,zhang2025gpr}. More recent studies have moved beyond conventional next-item prediction
toward more complex settings that require joint optimization across
candidates, tasks, or entire recommendation slates. RankGR~\cite{fu2026rankgr}
incorporates ranking-oriented preference optimization,
SIGMA~\cite{yu2026sigma} supports instruction-driven multi-task
recommendation, and HiGR~\cite{pang2025higr} addresses hierarchical slate
generation.

\subsection{Semantic Identifiers for Recommendation}
\label{sec:related_semantic_identifiers}

Semantic identifiers are a key component of identifier-based generative
recommendation, determining how items are represented in the model's discrete
output space.
Most SID construction methods first encode item content into
continuous embeddings and then discretize them through vector quantization.
VQ-VAE~\cite{van2017neural} integrates single-level vector quantization into
a neural autoencoding framework. Hierarchical codes can be obtained by
repeatedly quantizing the residual, either through classical residual vector
quantization (RVQ)~\cite{chen2010residual}, which sequentially applies
K-means, or through RQ-VAE~\cite{lee2022autoregressive}, which learns residual
codebooks within a neural autoencoder.

Recent SID studies address both the information encoded in identifiers and how those identifiers are learned. DAS~\cite{ye2025das} incorporates collaborative signals, while MMQ~\cite{xu2026mmq} combines multimodal information with behavior-aware adaptation. ETEGRec~\cite{liu2025etegrec} jointly trains the tokenizer and recommender, whereas UniSID~\cite{jiang2026endtoend} learns item embeddings and SIDs end to end. GRID~\cite{ju2025generativerecommendationsemanticids} and FORGE~\cite{fu2026forge} systematically compare SID construction choices on public benchmarks and industrial data.

\subsection{Preference and Reward Alignment for Generation}
\label{sec:related_preference_reward_alignment}

Preference and reward signals have been widely used to align language-model
generation with desired objectives. RLHF~\cite{ouyang2022training} optimizes
language models against learned human-preference rewards, while
DPO~\cite{rafailov2023direct} directly learns from preference pairs without
explicit reward modeling. GRPO~\cite{shao2024deepseekmath} further performs
group-relative policy optimization without a critic model. Beyond
outcome-level alignment, process reward models provide supervision for
intermediate reasoning steps~\cite{lightman2024verify} and can guide
inference-time search toward more promising trajectories
\cite{hung2025reward}.

Similar alignment techniques have been introduced into industrial generative
recommendation. OneRec~\cite{deng2025onerec} uses reinforcement learning to
align identifier generation with personalized preference signals.
RankGR~\cite{fu2026rankgr} uses listwise direct preference optimization to
align generation probabilities with ordered behavioral feedback, and trains
a binary target-codeword scorer to refine the candidates retained at each
decoding step.
PROMISE~\cite{guo2026promise} trains a path-level process reward model to
distinguish ground-truth SID prefixes from sampled valid prefixes, and uses
the resulting relevance scores to filter expanded intermediate candidates
during beam search, thereby mitigating semantic drift.

\section{Methodology}
\label{sec:method}

\subsection{Overview}
Figure~\ref{fig:framework} illustrates the overall framework of UAGR, which incorporates downstream utility into both SID construction and autoregressive decoding.
Following the preliminaries in Section~\ref{sec:preliminaries}, we first introduce Utility-Guided Decoding (Section~\ref{sec:utility_guided_decoding}), which estimates post-exposure outcomes and incorporates them into beam pruning, as illustrated in
Figures~\ref{fig:framework}(b) and~\ref{fig:framework}(c).
We then present Utility-Enriched Semantic Codes (Section~\ref{sec:utility_enriched_semantic_codes}), which preserve utility-relevant distinctions in the SID space, as shown in Figure~\ref{fig:framework}(a).

\subsection{Preliminaries}
\label{sec:preliminaries}

In a typical SID-based generative recommender, each item's multimodal embedding \(\mathbf{x}_i\in\mathbb{R}^{d}\), which encodes textual, visual, and other item-side features, is quantized into a hierarchical SID using residual K-means.
Starting from \(\mathbf{r}_i^{(0)}=\mathbf{x}_i\), the quantizer selects the nearest centroid and computes the updated residual \(\mathbf{r}_i^{(l)}\) at each level \(l\):

\begin{equation}
s_i^{(l)}
=
\arg\min_{k}
\left\|
\mathbf{r}_i^{(l-1)}-\mathbf{c}_{k}^{(l)}
\right\|_2^2,
\qquad
\mathbf{r}_i^{(l)}
=
\mathbf{r}_i^{(l-1)}
-
\mathbf{c}_{s_i^{(l)}}^{(l)},
\end{equation}
where $\mathbf{c}_{k}^{(l)}$ denotes the $k$-th centroid at level $l$.
Repeating this procedure over $L$ levels assigns each item a coarse-to-fine SID sequence:

\begin{equation}
\mathbf{s}_i
=
\left(
s_i^{(1)},s_i^{(2)},\ldots,s_i^{(L)}
\right).
\end{equation}

Given the interaction history $\mathcal{H}_u=(i_1,i_2,\ldots,i_n)$ of user $u$, a generative recommender autoregressively predicts the SID of a target item $i$, with the cumulative log-probability given by:

\begin{equation}
\log p_{\theta}
\left(
\mathbf{s}_i\mid\mathcal{H}_u
\right)
=
\sum_{l=1}^{L}
\log p_{\theta}
\left(
s_i^{(l)}
\mid
\mathbf{s}_i^{(<l)},\mathcal{H}_u
\right).
\end{equation}

During inference, beam search iteratively expands SID prefixes and retains the top-$K$ prefixes by cumulative generation log-likelihood at each step.

However, this likelihood-based pruning strategy does not explicitly account for downstream utility.
Candidates with similar generation probabilities can yield substantially different post-exposure outcomes, while SID prefixes of potentially high-utility items may be prematurely pruned during beam search.
To address this mismatch, we propose Utility-Guided Decoding (UGD), which combines generation log-likelihood with a utility score derived from predictions of multiple post-exposure outcomes to guide beam pruning.

\subsection{Utility-Guided Decoding (UGD)}
\label{sec:utility_guided_decoding}

\subsubsection{Utility Estimation Module.}

We consider $M$ post-exposure behavioral signals, such as watch time, long views, likes, and shares, with $m\in\{1,\ldots,M\}$ indexing each behavior.
As shown in Figure~\ref{fig:framework}(b), to estimate these signals for candidate SID prefixes, we reuse the hidden states of the generative backbone.
Specifically, at SID position $l$, given user history $\mathcal{H}_u$ and the generated prefix $\mathbf{s}^{(<l)}$, the decoder produces
\begin{equation}
\mathbf{h}^{(l)}
=
f_\theta\left(\mathcal{H}_u,\mathbf{s}^{(<l)}\right),
\end{equation}
where $f_\theta$ denotes the generative backbone. We then project the hidden state into a shared low-dimensional utility representation:
\begin{equation}
\mathbf{g}^{(l)}
=
P_\psi\left(\operatorname{sg}(\mathbf{h}^{(l)})\right),
\end{equation}
where $P_\psi$ is a projection network parameterized by $\psi$, implemented as a multilayer perceptron (MLP). The stop-gradient operator $\operatorname{sg}(\cdot)$ prevents utility-related gradients from updating the generative backbone, thereby avoiding interference from heterogeneous behavioral supervision and preserving its original generative capability.

Based on this shared representation, we introduce a separate token-wise prediction head for each behavior $m$. When expanding the current prefix with candidate token $v$ at SID position $l$, the corresponding head uses a learnable token embedding $\mathbf{e}_{m,v}^{(l)}$ and bias $b_{m,v}^{(l)}$ to compute the behavioral prediction score:
\begin{equation}
q_{m,v}^{(l)}
=
\left(\mathbf{e}_{m,v}^{(l)}\right)^\top
\mathbf{g}^{(l)}
+
b_{m,v}^{(l)}.
\end{equation}

\subsubsection{Utility-Guided Beam Search}
\label{sec:utility_guided_beam_search}

\begin{figure}[!t]
    \centering
    \includegraphics[
    width=\columnwidth,
    trim={0cm 1.05cm 0cm 0cm},
    clip
]{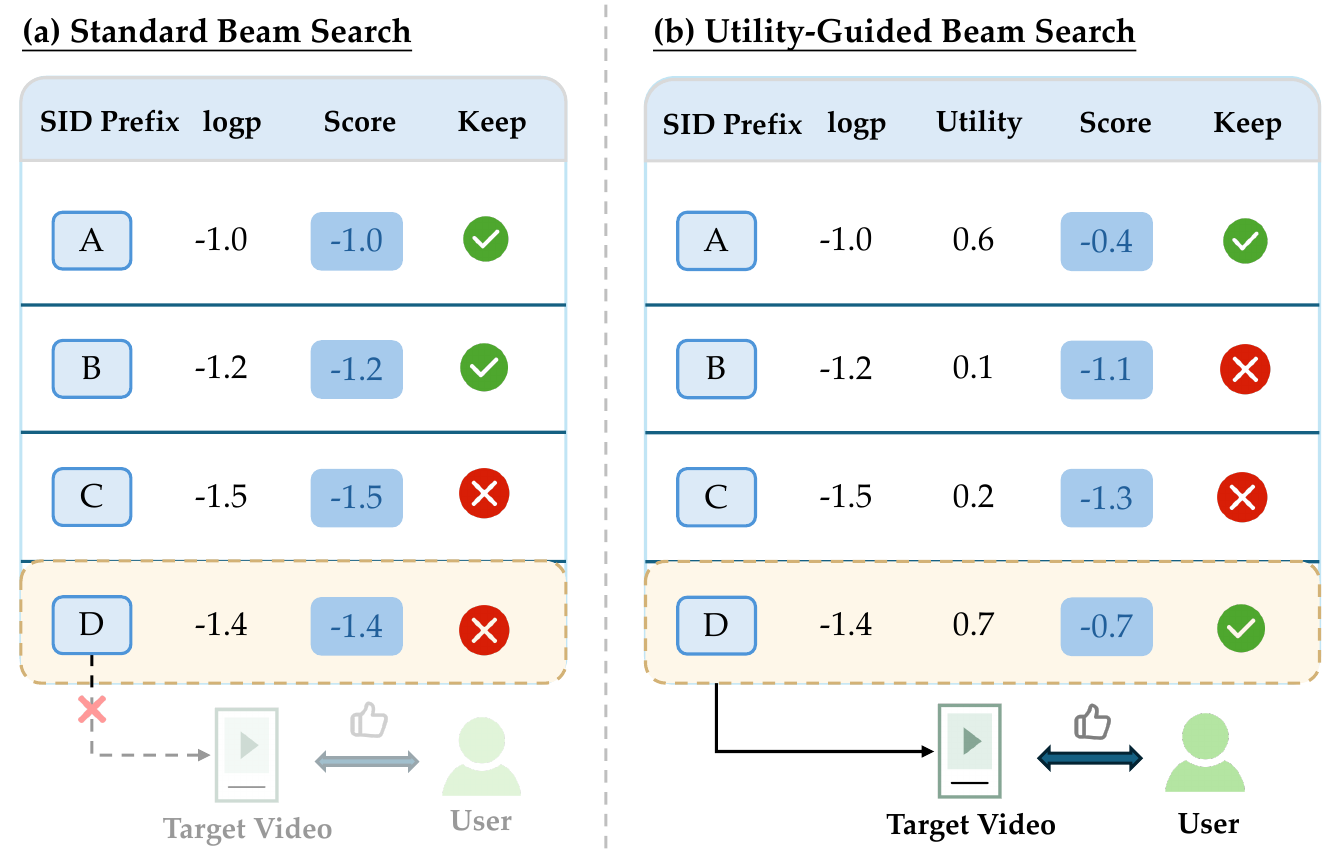}
    \caption{Illustration of utility-guided beam search.
    Expanded SID prefixes are scored by cumulative generation
    log-likelihood and current-step utility estimates before pruning to retain the top-$K$ prefixes.}
    \Description{At each decoding step, candidate SID prefixes are expanded.
    Their cumulative generation log-likelihood and current-step utility
    estimates jointly determine which prefixes survive pruning and continue
    to the next step.}
    \label{fig:beam_search}
\end{figure}

As shown in Figure~\ref{fig:framework}(c), we incorporate the estimated utility scores directly into beam pruning to reduce the risk of prematurely discarding high-utility candidates during beam search.
Figure~\ref{fig:beam_search} compares standard beam search with utility-guided beam search, illustrating how utility-aware scoring changes which candidate prefixes survive pruning.

Specifically, at SID position $l$, we compute an overall score for each prefix $\mathbf{s}^{(<l)}$ expanded with candidate token $v$ by combining generation likelihood and estimated utility:

\begin{equation}
S^{(l)}
=
\log p_{\theta}
\left(
\mathbf{s}^{(<l)},v
\mid
\mathcal{H}_u
\right)
+
\lambda
\sum_{m=1}^{M}
w_m
\operatorname{Norm}
\left(
q_{m,v}^{(l)}
\right).
\label{eq:utility_guided_beam_score}
\end{equation}

Here, the first term is the cumulative generation log-likelihood, and the second term aggregates the normalized utility estimates of the $M$ behavioral signals.
The operator $\operatorname{Norm}(\cdot)$ performs behavior-wise normalization to place heterogeneous behavioral predictions on a comparable scale before aggregation.
The coefficient $w_m$ specifies the contribution of behavior $m$, while $\lambda$ balances generation likelihood and downstream utility. The top-$K$ expanded prefixes under $S^{(l)}$ are retained for the next decoding step.

Unlike the generation likelihood, which is accumulated across SID positions, the utility score is applied only at the current position and is not carried forward to subsequent decoding steps.
This avoids repeatedly counting utility contributions across decoding steps while allowing utility to guide each pruning decision.

\subsubsection{Training Objectives}
\label{sec:training_objectives}

For each training sample, the target item is represented by an SID $\mathbf{s}=(s^{(1)},\ldots,s^{(L)})$ and associated with observed post-exposure feedback $\mathbf{y}=(y_1,\ldots,y_M)$, where $y_m$ denotes the feedback label for behavior $m\in\{1,\ldots,M\}$.
We use the observed feedback to supervise behavioral predictions at each position of the target SID.
Specifically, the utility-estimation loss for behavior $m$ at position $l$ is

\begin{equation}
\mathcal{L}_{m}^{(l)}
=
\ell_m
\left(
q_{m,s^{(l)}}^{(l)},
y_m
\right),
\label{eq:single_behavior_loss}
\end{equation}
where $\ell_m$ denotes the task-specific loss function, using cross-entropy for classification tasks and mean squared error for regression tasks.
Consequently, the utility loss is aggregated over $M$ behaviors and $L$ SID positions:

\begin{equation}
\mathcal{L}_{\mathrm{utility}}
=
\sum_{m=1}^{M}
\alpha_m
\sum_{l=1}^{L}
\mathcal{L}_{m}^{(l)},
\label{eq:utility_loss}
\end{equation}
where $\alpha_m$ controls the contribution of behavior $m$.

Combining with the standard next-token prediction objective $\mathcal{L}_{\mathrm{NTP}}$ for the generative backbone, we define the overall training loss as
\begin{equation}
\begin{gathered}
\mathcal{L}
=
\mathcal{L}_{\mathrm{NTP}}
+
\beta
\mathcal{L}_{\mathrm{utility}},
\\
\mathcal{L}_{\mathrm{NTP}}
=
-
\sum_{l=1}^{L}
\log p_{\theta}
\left(
s^{(l)}
\mid
\mathbf{s}^{(<l)},
\mathcal{H}_u
\right).
\end{gathered}
\label{eq:joint_loss}
\end{equation}

Here $\beta$ controls the contribution of utility estimation.

\noindent\textbf{Practical Advantages}
The proposed design is lightweight, stable, and extensible, making it well suited for low-risk deployment in large-scale recommendation systems.

\begin{itemize}[leftmargin=*, labelindent=0pt, labelsep=0.5em]
    \item \textbf{Low training and inference costs.}
    By reusing decoder hidden states and sharing a low-dimensional projection across behaviors, the module requires only lightweight behavior-specific token-wise heads.
    During inference, these heads compute utility scores from the shared representation alongside next-token predictions, without additional backbone forward passes, thereby minimizing computational and memory overhead.
    Thus, this design is well suited to the strict model-update and online-latency constraints of large-scale industrial recommendation.

    \item \textbf{Training stability.}
    The stop-gradient operation isolates utility supervision from the next-token prediction objective, ensuring that utility-related gradients update only the estimation module without affecting the generative backbone.
    This separation preserves the backbone's generation capability and improves training stability when behavioral labels are noisy, heterogeneous, or frequently revised.

    \item \textbf{Modular multi-objective extension.}
    New utility signals can be incorporated through additional behavior-specific heads and supervision losses, without modifying or retraining the generative backbone.
\end{itemize}

\subsection{Utility-Enriched Semantic Codes (UE-SID)}
\label{sec:utility_enriched_semantic_codes}

Although UGD incorporates predicted utility into beam pruning, its effectiveness remains constrained by the underlying SID space.
Conventional semantic tokenizers often face two limitations.
First, residual quantization methods such as RQ-KMeans and RQ-VAE often suffer from SID collisions \cite{rajput2023recommender,zhang2026reliable}, where multiple items share the same complete SID and become indistinguishable during generation, which substantially undermines the effectiveness of our UGD.
Second, conventional SID construction primarily relies on multimodal semantics without explicitly incorporating downstream utility.
Although collaborative signals can enrich item representations, their temporal variability creates a trade-off between stale SIDs and disruptive updates to the generator's learned token space \cite{Baikalov_2026}.
To address these limitations, we introduce Utility-Enriched Semantic Codes (UE-SID) by redesigning only the final quantization level.

\subsubsection{Utility-Enriched Codebook Construction}
\label{sec:utility_codebook_construction}

To reduce SID collisions, we replace the final-level nearest-centroid assignment in residual K-means (Section~\ref{sec:preliminaries}) with lookup-free quantization (LFQ), inspired by MAGVIT-v2~\cite{yu2023language}, while retaining the first $L-1$ quantization levels unchanged.
LFQ employs an implicit binary code space instead of explicit nearest-neighbor codebook lookup and uses entropy regularization to encourage diverse code utilization.
Specifically, a learnable linear projection $f_{\mathrm{proj}}$ maps the resulting residual $\mathbf{r}_i^{(L-1)}$ to a $B$-dimensional representation:

\begin{equation}
\mathbf{z}_i
=
f_{\mathrm{proj}}(\mathbf{r}_i^{(L-1)}),
\qquad
\mathbf{z}_i\in\mathbb{R}^{B}.
\label{eq:lfq_projection}
\end{equation}

Here, $B$ denotes the total bit budget of the final-level token and each dimension is independently binarized as

\begin{equation}
\widehat{z}_{i,j}
=
\begin{cases}
+1, & z_{i,j}>0,\\
-1, & z_{i,j}\leq 0,
\end{cases}
\qquad
j=1,\ldots,B.
\label{eq:lfq_quantization}
\end{equation}

We then map $+1/-1$ to $1/0$ to form a $B$-bit semantic code.

During training, we employ a straight-through estimator~\cite{bengio2013estimating} to propagate gradients through binary quantization.
For an LFQ training mini-batch $\mathcal{B}$, we define $p_i(c)$ as the temperature-scaled soft assignment of item $i$ to an implicit binary code $c$.
The entropy regularization objective is

\begin{equation}
\mathcal{L}_{\mathrm{ent}}
=
\frac{1}{|\mathcal{B}|}
\sum_{i\in\mathcal{B}}
H\left(p_i\right)
-
\gamma
H\left(
\frac{1}{|\mathcal{B}|}
\sum_{i\in\mathcal{B}}
p_i
\right),
\label{eq:lfq_entropy_loss}
\end{equation}
where the first term encourages confident assignments, while the second promotes diverse code utilization, weighted by $\gamma$.
We further introduce a commitment loss to constrain the projected representations toward their quantized values.
The overall LFQ training objective is

\begin{equation}
\begin{aligned}
\mathcal{L}_{\mathrm{LFQ}}
&=
\mathcal{L}_{\mathrm{com}}
+
\lambda_{\mathrm{ent}}
\mathcal{L}_{\mathrm{ent}},\\
\mathcal{L}_{\mathrm{com}}
&=
\frac{1}{|\mathcal{B}|}
\sum_{i\in\mathcal{B}}
\left\|
\mathbf{z}_i
-
\operatorname{sg}(\widehat{\mathbf{z}}_i)
\right\|_2^2,
\end{aligned}
\label{eq:lfq_loss}
\end{equation}
where $\operatorname{sg}(\cdot)$ denotes stop-gradient.

To incorporate item-level utility into semantic identifiers, we
discretize the utility signal into buckets and encode the bucket index
using $B_{\mathrm{u}}$ bits. Observing that some LFQ bits contribute little to SID distinguishability
(Appendix~\ref{app:weak_lfq}), we select and mask the subset of $B_{\mathrm{u}}$ bits whose removal causes the smallest increase in SID collisions.
The remaining $B_{\mathrm{s}}=B-B_{\mathrm{u}}$ bits are packed into the semantic code $c_{\mathrm{s},i}$, while the released bit capacity is allocated to utility encoding without increasing the final-level vocabulary size.

Let $a_i$ denote a utility statistic derived from historical behavioral feedback.
We discretize $a_i$ into $2^{B_{\mathrm{u}}}$ buckets and use its bucket index as the utility code:

\begin{equation}
c_{\mathrm{u},i}
=
\operatorname{Bucketize}
\left(
a_i
\right),
\qquad
c_{\mathrm{u},i}
\in
\left\{
0,\ldots,2^{B_{\mathrm{u}}}-1
\right\}.
\label{eq:utility_bucket}
\end{equation}

Formally, $\operatorname{Bucketize}(a_i)=\sum_{k=1}^{2^{B_{\mathrm{u}}}-1}\mathbb{I}[a_i\geq\tau_k]$, where $\tau_1<\cdots<\tau_{2^{B_{\mathrm{u}}}-1}$ are the ordered bucket boundaries and $\mathbb{I}[\cdot]$ denotes the indicator function.

Finally, we concatenate the semantic and utility codes to form the final-level SID token:

\begin{equation}
s_i^{(L)}
=
\operatorname{Concat}(c_{\mathrm{s},i},c_{\mathrm{u},i}),
\qquad B_{\mathrm{s}}+B_{\mathrm{u}}=B.
\label{eq:utility_enriched_code}
\end{equation}

\subsubsection{Online SID Assignment and Update}
\label{sec:online_sid_assignment_and_update}

For newly introduced items, we assign UE-SID through an online encoding pipeline: 
Given an item embedding $\mathbf{x}_i$, the pipeline first applies the fixed residual K-means centroids, the trained LFQ module, and the selected bit mask to obtain the first $L-1$ semantic tokens and the semantic code $c_{\mathrm{s},i}$.
It then retrieves the latest available behavioral statistics, computes the item-level utility statistic $a_i$, and obtains the utility code $c_{\mathrm{u},i}$ through bucketization.
The semantic and utility components are combined according to Eq.~\eqref{eq:utility_enriched_code}, and the resulting SID is published to the online code store.
This process requires only fixed-model inference, behavioral-feature lookup, bucket assignment, and bit composition, without retraining the semantic quantizers, thereby supporting low-latency online serving.

\begin{table*}[!t]
\centering
\caption{Overall offline performance on KuaiRand (all metrics in \%).
The best result is bold, and the second-best is underlined.}
\label{tab:overall_offline}
\large
\setlength{\tabcolsep}{13pt}
\renewcommand{\arraystretch}{1.05}
\begin{tabular}{llcccc}
\toprule
\multirow{2}{*}{Type}
& \multirow{2}{*}{Method}
& \multicolumn{4}{c}{KuaiRand} \\
\cmidrule(l){3-6}
&
& Recall@3 $\uparrow$
& NDCG@3 $\uparrow$
& Recall@5 $\uparrow$
& NDCG@5 $\uparrow$ \\
\midrule

\multirow{5}{*}{Discriminative}
& Caser~\cite{tang2018caser}
& 2.0462 & 1.5378 & 3.0924 & 1.9753 \\
& SASRec~\cite{kang2018sasrec}
& 1.8460 & 1.4063 & 2.8637 & 1.8436 \\
& BERT4Rec~\cite{sun2019bert4rec}
& 1.6482 & 1.3563 & 2.6452 & 1.7629 \\
& S3-Rec~\cite{zhou2020s3rec}
& 2.3681 & 1.8630 & 3.4612 & 2.2975 \\
& HSTU~\cite{zhai2024hstu}
& 2.4831 & 1.8742 & 3.6223 & 2.3593 \\
\midrule

Unified
& LIGER~\cite{rajput2024liger}
& 1.9951 & 1.4632 & 3.0073 & 1.8872 \\
\midrule

\multirow{5}{*}{Generative}
& TIGER~\cite{rajput2023recommender}
& 2.0105 & 1.4941 & 2.9983 & 1.8999 \\
& RecForest~\cite{feng2022recforest}
& \underline{2.5748}
& \underline{1.9516}
& \underline{3.6472}
& \underline{2.4063} \\
& EAGER~\cite{wang2024eager}
& 2.5631 & 1.9480 & 3.5974 & 2.3810 \\
& OneRec-1.7B~\cite{wang2025openonerec}
& 2.1114 & 1.6432 & 3.1493 & 2.0460 \\
\cmidrule(l){2-6}
& \textbf{UAGR}
& \textbf{2.7741}
& \textbf{2.2740}
& \textbf{3.9632}
& \textbf{2.7764} \\
\bottomrule
\end{tabular}
\end{table*}

To reflect evolving item utility, we periodically refresh the utility component for items with updated behavioral statistics.
Every hour, the update pipeline recomputes $a_i$ from the latest statistics and updates $c_{\mathrm{u},i}$ accordingly, while keeping the first $L-1$ tokens and the LFQ semantic component $c_{\mathrm{s},i}$ fixed.
This incremental update keeps the utility information fresh without perturbing the item's semantic representation.

Changes in item utility may also gradually shift the overall distribution of utility statistics.
We therefore recalibrate the bucket boundaries weekly based on the aggregated distribution of item-level utility statistics.
This two-timescale strategy balances freshness and stability: hourly updates capture recent changes in individual item utility, while weekly boundary adjustments accommodate gradual distribution shifts without frequently redefining utility levels.

\section{Experiments}
\label{sec:experiments}

In this section, we conduct offline and online experiments to answer
the following research questions:
\begin{itemize}[leftmargin=*, labelindent=0pt, labelsep=0.5em]
    \item \textbf{RQ1:} How does UAGR perform compared with existing recommendation methods?
    \item \textbf{RQ2:} How does UE-SID construction compare with existing methods in codebook quality and downstream recommendation performance?
    \item \textbf{RQ3:} How does each component affect the performance of UAGR?
    \item \textbf{RQ4:} How does UGD compare with different utility alignment strategies?
    \item \textbf{RQ5:} How does UAGR perform in a real-world online recommendation scenario?
\end{itemize}

\subsection{Experimental Setup}
\label{sec:experimental_setup}

\subsubsection{Baselines}
To evaluate the overall recommendation effectiveness of UAGR, we compare it with discriminative methods, including Caser~\cite{tang2018caser}, SASRec~\cite{kang2018sasrec}, BERT4Rec~\cite{sun2019bert4rec}, S3-Rec~\cite{zhou2020s3rec}, and HSTU~\cite{zhai2024hstu}; generative methods, including TIGER~\cite{rajput2023recommender}, RecForest~\cite{feng2022recforest}, EAGER~\cite{wang2024eager}, and OneRec-1.7B~\cite{wang2025openonerec}; and a unified method, LIGER~\cite{rajput2024liger}.
To assess the effectiveness of the proposed UE-SID component, we compare it with existing SID construction methods, including RQ-KMeans~\cite{chen2010residual}, OPQ~\cite{ge2013optimized}, RQ-OPQ~\cite{chen2025onesearch}, LCRec~\cite{zheng2024lcrec}, FORGE~\cite{fu2026forge}, LETTER~\cite{wang2024letter}, and UIST-RQ~\cite{liu2024discrete}.
Detailed baseline descriptions are provided in Appendix~\ref{app:baselines}.

\subsubsection{Dataset and Metrics}
We use KuaiRand~\cite{gao2022kuairand}, a public recommendation dataset collected on Kuaishou that provides multiple types of user feedback.
Further dataset details are provided in Appendix~\ref{app:dataset}.
We evaluate recommendation performance using Recall and NDCG at cutoffs 3 and 5.
To evaluate codebook construction, we report Independent Coding Rate (ICR)~\cite{chen2025onesearch}, Gini coefficient~\cite{fu2026forge}, and code entropy~\cite{kuai2024breakinghourglassphenomenonresidual}.
ICR measures the proportion of items assigned unique, collision-free SIDs, while lower Gini and higher entropy indicate more balanced code utilization.

\subsubsection{Implementation Details}
For data splitting, we adopt the widely used leave-one-out protocol. We adopt OneRec-1.7B as the generative backbone and apply low-rank adaptation (LoRA)~\cite{hu2022lora} for parameter-efficient fine-tuning.
The model is trained for 4 epochs on NVIDIA L20 GPUs with a batch size of 64. We employ a cosine learning rate schedule with peak and minimum learning rates of $2 \times 10^{-5}$ and $1 \times 10^{-5}$, respectively, and clip the gradient norm at 1.0.
The SID configuration is $3 \times 8192$ (three levels with 8,192 codes each), with a beam size of 200.
For UE-SID, the final level uses $B_{\mathrm{s}}=10$ bits for semantic encoding and $B_{\mathrm{u}}=3$ bits for utility encoding.
Further implementation details are provided in Appendix~\ref{app:implementation}.

\begin{table*}[!t]
\centering
\caption{Comparison of different SID construction methods on KuaiRand
(Recall and NDCG in \%).
The best result is highlighted in bold, and the second-best
result is underlined.}
\label{tab:sid_comparison}
\large
\setlength{\tabcolsep}{9pt}
\renewcommand{\arraystretch}{1.1}
\begin{tabular}{lccccccc}
\toprule
\multirow{2}{*}{Method}
& \multicolumn{7}{c}{KuaiRand} \\
\cmidrule(l){2-8}
& Recall@3 $\uparrow$
& NDCG@3 $\uparrow$
& Recall@5 $\uparrow$
& NDCG@5 $\uparrow$
& ICR $\uparrow$
& Gini $\downarrow$
& Entropy $\uparrow$ \\
\midrule

RQ-KMeans~\cite{chen2010residual}
& 2.1114 & 1.6432 & 3.1493 & 2.0460
& 0.9571 & 0.02660 & 5.9732 \\

OPQ~\cite{ge2013optimized}
& 2.0095 & 1.6174 & 3.0695 & 2.0323
& 0.9582 & \underline{0.00941} & 7.9522 \\

RQ-OPQ~\cite{chen2025onesearch}
& \underline{2.2364} & \underline{1.8171}
& 3.2123 & 2.2171
& 0.9843 & 0.01030 & 7.8756 \\

UIST-RQ~\cite{liu2024discrete}
& 2.2137 & 1.7982 & 3.1641 & 2.1943
& 0.9842 & 0.00990 & 7.8642 \\

LCRec~\cite{zheng2024lcrec}
& 2.1473 & 1.7463
& \underline{3.3651} & \underline{2.2579}
& \textbf{0.9997} & 0.01250 & 7.5144 \\

FORGE~\cite{fu2026forge}
& 2.0744 & 1.4785 & 2.9784 & 1.8563
& 0.9687 & 0.01760 & 7.5762 \\

LETTER~\cite{wang2024letter}
& 2.1462 & 1.6853 & 3.0845 & 2.0753
& 0.9823 & 0.00980 & \underline{8.0147} \\

\midrule
\textbf{UE-SID}
& \textbf{2.4089} & \textbf{1.9669}
& \textbf{3.5851} & \textbf{2.4357}
& \underline{0.9902}
& \textbf{0.00500} & \textbf{8.0918} \\
\bottomrule
\end{tabular}
\end{table*}

\subsection{Overall Offline Evaluation (RQ1)}
\label{sec:offline_evaluation}

Table~\ref{tab:overall_offline} reports the offline performance of UAGR and the baselines on KuaiRand.
UAGR achieves the best performance across all evaluation metrics, outperforming the strongest baseline by 7.74\% in Recall@3 and 16.52\% in NDCG@3.
The improvement in Recall@$K$ indicates that UAGR retrieves more items associated with positive user feedback, while the improvement in NDCG@$K$ shows that these items are ranked higher in the recommendation list.
These results demonstrate the effectiveness of UAGR in improving recommendation performance.

Specifically, UAGR outperforms all evaluated generative baselines, including methods that incorporate collaborative signals beyond semantic information.
For example, although EAGER~\cite{wang2024eager} integrates semantic and collaborative signals through a two-stream architecture, UAGR consistently outperforms EAGER across all evaluated metrics, highlighting the effectiveness of our explicit, stability-preserving utility integration compared with EAGER's collaborative-signal fusion approach.
UE-SID retains utility-relevant distinctions in the discrete identifier space, while UGD incorporates predicted utility into beam pruning to reduce the premature removal of high-utility candidates.
Together, these components align item tokenization and generation with downstream utility, contributing to UAGR's superior performance on KuaiRand.



\subsection{Evaluation of Utility-Enriched Semantic Codes (RQ2)}
\label{sec:sid_evaluation}

To evaluate our SID construction method, we compare UE-SID with alternative construction methods on KuaiRand.
For a fair comparison, all methods use the same OneRec-1.7B backbone for downstream generative recommendation.
The results are presented in Table~\ref{tab:sid_comparison}.
UE-SID achieves strong results in both codebook construction quality and downstream recommendation performance.

Regarding SID construction quality, UE-SID achieves a high ICR, along with the lowest Gini coefficient and the highest code entropy among all compared methods.
The high ICR benefits from LFQ and the replacement of weakly discriminative semantic bits with utility codes, which further distinguish items with similar semantic representations.
In particular, utility-based bucketing further contributes to a reduction in SID collisions, as detailed in Appendix~\ref{app:eff_util}.
Meanwhile, the entropy regularization objective in Eq.~\eqref{eq:lfq_entropy_loss} discourages code collapse and promotes balanced code utilization, contributing to the favorable Gini and entropy results.

Regarding downstream recommendation performance, UE-SID achieves the best results across all evaluated metrics.
Several baselines, including LCRec~\cite{zheng2024lcrec}, FORGE~\cite{fu2026forge}, and LETTER~\cite{wang2024letter}, also explore incorporating collaborative signals into SID construction.
Specifically, LCRec learns quantized item indices and integrates collaborative semantics through language-model alignment tasks; FORGE incorporates item co-occurrence signals through contrastive learning before quantization; and LETTER aligns quantized item representations with collaborative-filtering embeddings through contrastive regularization.
In contrast, UE-SID explicitly discretizes item-level utility into buckets and incorporates the resulting utility codes into the final-level token of each SID.
This design enables stable utility integration without altering the learned semantic codes, thereby outperforming these collaborative-signal-enhanced methods across all recommendation metrics.
Notably, although LCRec achieves the highest Independent Coding Rate, its downstream recommendation performance remains below that of UE-SID, suggesting that SID uniqueness alone does not guarantee superior recommendation performance.

\subsection{Ablation Study (RQ3)}
\label{sec:ablation}

To assess the contribution of each proposed component in UAGR, we conduct an ablation study on KuaiRand.
We add each component independently to the same OneRec-1.7B base model and evaluate its effect on downstream recommendation performance.
We additionally report ICR for the codebook construction variants to assess their effect on SID distinguishability.
The results are reported in Table~\ref{tab:ablation}.

The results show that each proposed component improves downstream recommendation performance when independently added to the same base model.
Introducing LFQ alone increases ICR and improves recommendation performance, as it replaces nearest-centroid assignment with an implicit binary code space and encourages more balanced code utilization through entropy regularization, thereby reducing SID collisions.
Incorporating utility codes further increases ICR and improves recommendation performance beyond LFQ alone.
By distinguishing semantically similar items according to their utility levels, UE-SID further reduces collisions and preserves distinctions relevant to downstream recommendation.

Notably, Utility-Guided Decoding achieves the largest improvements in recommendation performance among the individual variants.
These results suggest that likelihood-based next-item prediction alone is insufficient to fully address downstream recommendation objectives.
UGD incorporates predicted utility into decoding, and its superior performance highlights the value of utility information as a complementary signal for retaining high-utility candidates during decoding.

\begin{figure}[!t]
    \centering
    \includegraphics[width=\columnwidth]{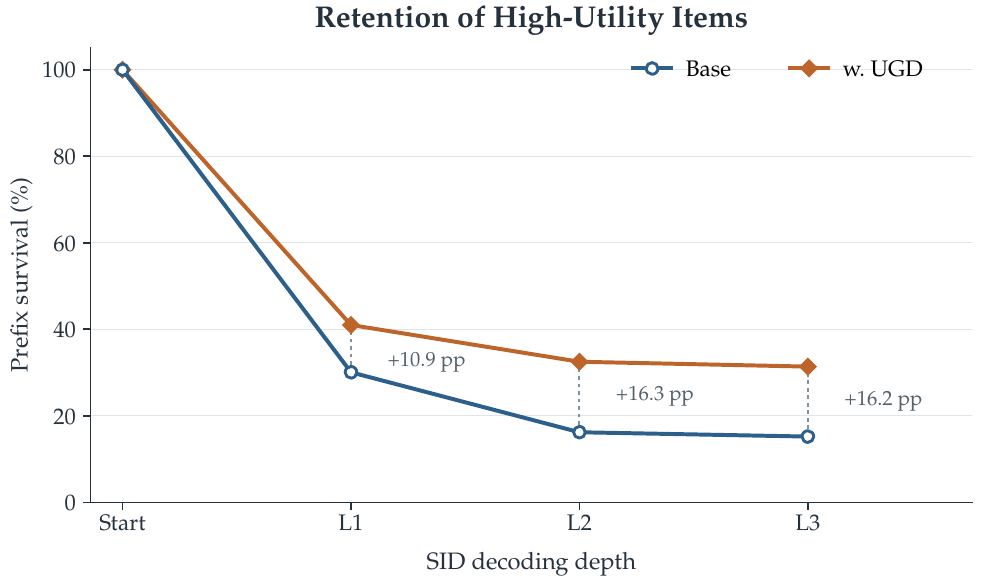}
    \caption{High-utility prefix retention during beam search for Base and
Base with UGD at the same beam width. Annotated gaps show improvements
in percentage points.}
    \Description{The figure compares high-utility prefix retention during
    beam search between Base and Base with UGD at the same beam width.
    Gap annotations show differences in retention rates in percentage points.}
    \label{fig:high_utility_retention}
\end{figure}

To better illustrate how UGD reduces premature pruning, we track the survival of high-utility candidates, defined as those with positive long-view feedback (\texttt{long\_view}=1), during beam search.
At each SID decoding position, we measure the proportion of these items whose corresponding prefixes survive beam pruning, using the initial high-utility item set as the fixed denominator.
Both variants use the same beam width.
As shown in Figure~\ref{fig:high_utility_retention}, UGD retains more high-utility candidates than the baseline, with the largest improvement observed at the second decoding level.
These results demonstrate that incorporating utility information into beam search reduces the premature pruning of high-utility items, further validating the effectiveness of our UGD.

\begin{table}[!t]
\centering
\caption{Component ablations and comparisons of other utility alignment strategies
on KuaiRand (Recall and NDCG in \%). Each individual variant is independently added to Base.}
\label{tab:ablation}
\normalsize
\renewcommand{\arraystretch}{1.05}
\setlength{\tabcolsep}{6pt}
\begin{tabular}{lccccc}
\toprule
\multirow{2}{*}{Variant}
& \multicolumn{2}{c}{Recall@$K$}
& \multicolumn{2}{c}{NDCG@$K$}
& \multirow{2}{*}{ICR} \\
\cmidrule(lr){2-3}
\cmidrule(lr){4-5}
& $K=3$ & $K=5$ & $K=3$ & $K=5$ & \\
\midrule
\textbf{Full UAGR}
& \textbf{2.774}
& \textbf{3.963}
& \textbf{2.274}
& \textbf{2.776}
& \textbf{0.9902} \\

\midrule
Base
& 2.111 & 3.149 & 1.643 & 2.046 & 0.9571 \\

w. LFQ
& 2.246 & 3.288 & 1.654 & 2.064 & 0.9789 \\

w. UE-SID
& 2.409 & 3.585 & 1.967 & 2.436 & 0.9902 \\

w. UGD
& 2.578 & 3.712 & 2.047 & 2.499 & -- \\

\midrule
w. Reranking
& 2.307 & 3.275 & 1.675 & 2.083 & -- \\

w. GRPO
& 2.418 & 3.602 & 1.953 & 2.450 & -- \\
\bottomrule
\end{tabular}
\end{table}

\begin{table}[!t]
\centering
\caption{Online A/B test results on the Kuaishou platform.
All reported user-consumption gains are statistically significant
under a two-sided $t$-test with $p<0.05$.}
\label{tab:online_ab}
\small
\setlength{\tabcolsep}{4pt}
\renewcommand{\arraystretch}{1.2}
\begin{tabular}{lcccc}
\toprule
& App Usage Time & Video Watch Time & Video Views & QPS \\
\midrule
$\Delta$
& $+0.074\%$ & $+0.206\%$ & $+0.164\%$ & $-0.03\%$ \\
\bottomrule
\end{tabular}
\end{table}

\subsection{Comparison of Utility Alignment Strategies (RQ4)}
\label{sec:utility_alignment_comparison}

To more comprehensively evaluate our method, we compare strategies that incorporate utility information at different stages of generative recommendation.
Specifically, we consider training-time utility alignment through reinforcement learning, implemented using GRPO~\cite{shao2024deepseekmath}, and post-generation reranking, which reorders generated candidates according to their predicted utility.
Both strategies are applied to the same OneRec-1.7B base model. For a fair comparison, GRPO rewards and reranking scores use the same utility objectives
and corresponding weights as our method, with detailed configurations
provided in Appendix~\ref{app:implementation}.

As shown in Table~\ref{tab:ablation}, UAGR outperforms both GRPO and post-generation reranking across all evaluation metrics.
In particular, Base with UGD alone also outperforms GRPO.
Compared with Base, GRPO improves recommendation performance,
demonstrating the effectiveness of reward-based training.
However, GRPO optimizes rewards over complete SID sequences, whereas beam search makes irreversible pruning decisions over intermediate SID prefixes.
Such sequence-level alignment does not necessarily ensure that high-utility candidates survive intermediate pruning.
In contrast, UGD explicitly estimates candidate utility at each SID decoding level and incorporates it into the corresponding beam-pruning scores, directly guiding prefix selection throughout generation.
UGD also outperforms post-generation reranking, which can only reorder retained candidates and cannot recover high-utility items whose paths have already been pruned.
These comparisons demonstrate the effectiveness of our method
and support our design choices.

\subsection{Online A/B Test Results~(RQ5)}
\label{sec:online_ab}

We deploy UAGR in the short-video recommendation scenario on the Kuaishou platform, replacing the production base model, and conduct a one-week online A/B test to evaluate its effectiveness in a large-scale recommendation environment.
The reported improvements in user-consumption metrics are statistically significant under a two-sided $t$-test with $p<0.05$.
We additionally report the change in serving throughput, measured by queries per second (QPS).
The results are presented in Table~\ref{tab:online_ab}.

Compared with the production baseline, UAGR improves app usage time by
0.074\%, video watch time by 0.206\%, and video views by 0.164\%, with
only a 0.03\% reduction in QPS.
These online results demonstrate the effectiveness of UAGR in a large-scale production recommendation system and further highlight the value of explicitly incorporating downstream utility into generative recommendation. Further demonstrating its practical viability, UAGR has been fully deployed in Kuaishou's production recommendation system since September 2026.


\section{Conclusion}
In this paper, we proposed UAGR to incorporate downstream utility into both SID construction and autoregressive decoding.
Utility-Enriched Semantic Codes preserve utility-relevant item distinctions without disrupting the underlying hierarchical semantic structure or increasing SID length and vocabulary size.
It also supports incremental utility updates without retraining
the semantic quantizers.
Utility-Guided Decoding incorporates predicted post-exposure utility into beam scoring to reduce premature pruning of high-utility candidates, without interfering with generative backbone training.
It also enables flexible utility control through configurable
weights for different behavioral objectives.
Extensive offline experiments and online A/B tests demonstrate
the effectiveness of UAGR and support its design choices.
UAGR has been fully deployed on Kuaishou.
These findings highlight the importance of explicit utility modeling at both the representation and decoding stages of generative recommendation, and we hope they will inspire further research in this direction.



\bibliographystyle{ACM-Reference-Format}
\bibliography{references}

\appendix

\section{Dataset Description}
\label{app:dataset}

KuaiRand~\cite{gao2022kuairand} is a public short-video recommendation
dataset collected from real user interaction logs on the Kuaishou
platform. To mitigate exposure bias, it introduces randomized exposure
into normal recommendation feeds by replacing selected recommendations
with videos uniformly sampled from a pool of 7,583 items. The dataset
contains standard recommendation logs from 15 recommendation policies
and randomized-exposure logs. It provides
user and video identifiers, interaction timestamps, 30 user features,
62 video features, and 12 feedback-related signals, including clicks
or effective plays, likes, follows, comments, shares, long views, and
viewing duration. These heterogeneous behavioral signals support
the modeling of post-exposure utility.

KuaiRand is released in three versions. KuaiRand-27K contains the
complete released logs for 27,285 users, preserving their large-scale
interaction histories. KuaiRand-1K uniformly samples 1,000 users from
KuaiRand-27K and retains their interaction logs and associated videos.
KuaiRand-Pure retains only interactions involving videos in the
randomized-exposure candidate pool, resulting in a smaller item space
but incomplete user interaction histories. We use KuaiRand-27K for
our offline experiments. Table~\ref{tab:dataset_statistics} summarizes
the original release statistics of the three versions before
experiment-specific preprocessing.

\begin{table}[H]
\centering
\caption{Statistics of KuaiRand before preprocessing.
Standard and Random denote standard recommendation and randomized
exposure, respectively.}
\label{tab:dataset_statistics}
\normalsize
\setlength{\tabcolsep}{2pt}
\renewcommand{\arraystretch}{1.15}
\begin{tabular}{@{}llrrr@{}}
\toprule
Dataset & Policy & Users & Videos & Interactions \\
\midrule
\multirow{2}{*}{KuaiRand-27K}
& Standard & 27,285 & 32,038,725 & 322,278,385 \\
& Random & 27,285 & 7,583 & 1,186,059 \\

\midrule
\multirow{2}{*}{KuaiRand-1K}
& Standard & 1,000 & 4,369,953 & 11,713,045 \\
& Random & 1,000 & 7,388 & 43,028 \\

\midrule
\multirow{2}{*}{KuaiRand-Pure}
& Standard & 27,285 & 7,551 & 1,436,609 \\
& Random & 27,285 & 7,583 & 1,186,059 \\
\bottomrule
\end{tabular}
\end{table}

\section{Detailed Experimental Settings}
\label{app:settings}

\subsection{Baselines}
\label{app:baselines}

\begin{table*}[!t]
\centering
\caption{Detailed ablation and comparative studies of UE-SID
on KuaiRand (Recall and NDCG in \%).
The best results are highlighted in bold, and the
second-best results are underlined.}
\label{tab:uesc_detailed_ablation}
\setlength{\tabcolsep}{9pt}
\renewcommand{\arraystretch}{1.1}
\begin{tabular}{lccccc}
\toprule
Method
& Collision (\%) $\downarrow$
& Recall@3 $\uparrow$
& NDCG@3 $\uparrow$
& Recall@5 $\uparrow$
& NDCG@5 $\uparrow$ \\
\midrule
Base (RQ-KMeans)
& 4.2915 & 2.1114 & 1.6432 & 3.1493 & 2.0460 \\

LFQ-13
& 2.1122 & 2.2461 & 1.6538 & 3.2875 & 2.0643 \\

LFQ-13 + Mask
& 2.6793
& \underline{2.3074}
& \underline{1.7147}
& \underline{3.2964}
& \underline{2.1117} \\

LFQ-13 + Mask + Utility (UE-SID)
& \textbf{0.9801}
& \textbf{2.4089}
& \textbf{1.9669}
& \textbf{3.5851}
& \textbf{2.4357} \\

LFQ-10 + Utility
& \underline{0.9967}
& 2.1675 & 1.6419 & 3.1972 & 2.0626 \\
\bottomrule
\end{tabular}
\end{table*}

We compare UAGR with representative discriminative, unified, and
generative recommendation methods.

\begin{itemize}[leftmargin=*, labelindent=0pt, labelsep=0.5em]
\item \textbf{Discriminative methods.}
Caser~\cite{tang2018caser} uses convolutional sequence
embedding to capture local patterns in user behavior.
SASRec~\cite{kang2018sasrec} models sequential dependencies
using unidirectional self-attention.
BERT4Rec~\cite{sun2019bert4rec} learns bidirectional sequence
representations through masked-item prediction.
S3-Rec~\cite{zhou2020s3rec} enhances sequential recommendation
through self-supervised pre-training.
HSTU~\cite{zhai2024hstu} introduces a scalable sequential
transducer for large-scale recommendation.

\item \textbf{Unified methods.}
LIGER~\cite{rajput2024liger} unifies generative retrieval
and dense retrieval in a single sequential recommendation framework.

\item \textbf{Generative methods.}
TIGER~\cite{rajput2023recommender} autoregressively generates
semantic identifiers of target items.
RecForest~\cite{feng2022recforest} represents items as paths
in learned tree-structured indexes and retrieves them through
hierarchical routing.
EAGER~\cite{wang2024eager} combines behavioral and semantic
streams for generative recommendation.
OneRec-1.7B from OpenOneRec~\cite{wang2025openonerec} serves as
the generative baseline and autoregressively generates item identifiers.
\end{itemize}

For SID construction, we compare UE-SID with the following methods.

\begin{itemize}[leftmargin=*, labelindent=0pt, labelsep=0.5em]
\item \textbf{Reconstruction-based methods.}
RQ-KMeans~\cite{chen2010residual} learns successive codebooks by
applying K-means to the residuals from preceding quantization levels.
OPQ~\cite{ge2013optimized} learns a rotation of the embedding space
to reduce product quantization error.
RQ-OPQ~\cite{chen2025onesearch} combines residual quantization
with optimized product quantization to construct discrete item codes.

\item \textbf{Collaborative-signal-enhanced methods.}
LCRec~\cite{zheng2024lcrec} aligns semantic identifiers with
collaborative information through auxiliary training objectives.
FORGE~\cite{fu2026forge} incorporates interaction-derived signals
into semantic identifier construction for industrial generative retrieval.
LETTER~\cite{wang2024letter} learns item tokenization by incorporating
collaborative signals and regularizing code utilization.

\item \textbf{CTR-supervised methods.}
UIST-RQ~\cite{liu2024discrete} fine-tunes item embeddings through
a CTR prediction task and applies residual quantization to obtain
discrete semantic codes.
\end{itemize}

\subsection{Implementation Details}
\label{app:implementation}

\paragraph{Data Splitting.}
We chronologically sort each user's interactions and construct
positive-feedback sequences using long-view interactions
(\texttt{long\_view}=1).
For each user, the last and second-to-last positive interactions
are held out for testing and validation, respectively, while the
remaining interactions are used for training.
To prevent information leakage, item-level utility statistics and supervision for the utility prediction heads are derived exclusively from the full training-stage exposure logs, including both long-view and non-long-view exposures.

\paragraph{Model and Training Configuration.}
The backbone is fine-tuned using LoRA with a rank of 16, a scaling
factor of 32, and a dropout rate of 0.05.
The model is loaded in BF16 precision with scaled dot-product
attention (SDPA).
Item representations are derived from 256-dimensional text embeddings
encoded by Qwen3-Embedding-4B~\cite{zhang2025qwen3embedding}.
The LFQ module is trained for 15 epochs using AdamW~\cite{loshchilov2019decoupled} with a batch size
of 256 and a learning rate of $1 \times 10^{-3}$.
The inverse temperature for LFQ is set to 10, with
$\lambda_{\mathrm{ent}}=0.01$.

For UE-SID, we use each item's empirical click-through rate (CTR)
as the utility statistic $a_i$ for bucket assignment.
For UGD, we use four utility signals: watch time, long view,
like, and share, with decoding weights $w_1=0.6$, $w_2=0.2$,
$w_3=0.1$, and $w_4=0.1$, respectively.
The normalization operator Norm in Eq. (7) applies sigmoid to the predicted logits of binary classification heads. Regression outputs, such as predicted watch time, are first standardized within the candidate set at each beam-search step and then passed through sigmoid.
The behavior-specific training weights in Eq.~\eqref{eq:utility_loss}
are set to $\alpha_m=1$ for all four signals.
The UGD coefficient is set to $\lambda=1$,
and the utility loss weight is set to $\beta=0.001$.

For the comparisons in Section~\ref{sec:utility_alignment_comparison},
GRPO and post-generation reranking use utility scores computed
by the utility estimation module with the same four behavioral
objectives and aggregation weights specified above.
The aggregated utility score serves as the reward for complete
SID sequences in GRPO and as the ranking criterion for generated
candidates in post-generation reranking.


\section{Detailed Ablation Studies on UE-SID}
\label{app:detailed_ablation_uesc}

\subsection{Experimental Setup and Bit Selection}
To further examine the design choices of UE-SID, we conduct detailed
ablation studies on KuaiRand, as reported in
Table~\ref{tab:uesc_detailed_ablation}.
All variants retain the first two RQ-KMeans levels and modify only
the final-level encoding, using the same OneRec-1.7B backbone
and training settings.
Specifically, \textbf{LFQ-13} replaces the final-level RQ-KMeans
with LFQ to produce a 13-bit semantic code.
\textbf{LFQ-13 + Mask} removes the three least discriminative bits,
retaining the remaining 10 semantic bits.
\textbf{LFQ-13 + Mask + Utility (UE-SID)} further appends three
utility bits, encoding item-level utility into eight discrete
buckets based on historical behavioral feedback.
For comparison, \textbf{LFQ-10 + Utility} directly learns a
10-bit LFQ semantic code and appends the same three utility bits.

To identify the least discriminative bits, we exhaustively
evaluate all $\binom{13}{3}=286$ three-bit combinations.
For each combination, we mask the selected bits and measure
the resulting increase in SID collision rate.
The combination yielding the smallest increase is selected,
minimizing the loss of item distinguishability.

\subsection{Effect of LFQ}
Replacing RQ-KMeans with LFQ substantially reduces SID collisions
while improving downstream recommendation performance.
Originally introduced in MAGVIT-v2~\cite{yu2023language}, LFQ addresses the difficulty
of learning and effectively utilizing large VQ codebooks, where
improved reconstruction quality does not necessarily translate
into better generation.
By replacing explicit codebook lookup with independent binary
decisions and encouraging balanced code utilization through
entropy regularization, LFQ supports more scalable discrete
representations.
Our results further demonstrate its effectiveness in improving
item distinguishability for generative recommendation.

\subsection{Effect of Masking Weakly Discriminative Bits}
\label{app:weak_lfq}
Masking the three least discriminative bits slightly increases
the collision rate but yields marginal improvements in
recommendation performance.
This suggests that greater code distinguishability does not
necessarily imply better downstream performance.
Some weakly discriminative bits may introduce fine-grained
partitions with limited relevance to recommendation, potentially
increasing token sparsity and learning difficulty.
Removing these bits may therefore act as a form of representation
regularization while preserving the most informative
semantic distinctions.

\subsection{Effect of Utility Injection}
\label{app:eff_util}
Appending three utility bits to the retained semantic code
substantially reduces collisions and consistently improves
all recommendation metrics.
By assigning items to utility buckets based on historical
behavioral feedback, UE-SID further distinguishes semantically
similar items with different downstream utility.
These results demonstrate that selectively reallocating
code capacity from weak semantic distinctions to utility
information improves both item distinguishability and
downstream recommendation quality.

\subsection{Comparison with Direct 10-Bit LFQ}
Finally, we compare UE-SID with directly learning a 10-bit
LFQ semantic code followed by three utility bits.
Despite their similar collision rates, UE-SID achieves
substantially better recommendation performance.
This indicates that identical code capacity and comparable
collision rates do not guarantee equivalent representation quality.
Directly restricting LFQ to 10 bits may impose an excessive
information bottleneck during quantization, discarding
semantic distinctions useful for recommendation.
In contrast, UE-SID first learns a richer 13-bit representation
and then selectively removes the least discriminative bits,
better preserving informative semantic structure.

These findings highlight the importance of informed capacity
allocation between semantic and utility information,
rather than simply concatenating independently learned codes.

\end{document}